\documentclass[preprint,12pt]{elsarticle}

\usepackage{amssymb}
\usepackage{amssymb,amsmath}
\usepackage{wrapfig}
\usepackage{graphicx,grffile}
\usepackage{lscape}
\usepackage{natbib}
\usepackage{adjustbox}
\usepackage{multirow}
\usepackage[T1]{fontenc} 
\usepackage{booktabs}
\usepackage{xurl}
\usepackage{xcolor}

\journal{Journal of Biomedical Informatics}

\begin{document}

\begin{frontmatter}



\title{DR.BENCH: Diagnostic Reasoning Benchmark for Clinical Natural Language Processing}


\author[inst1]{Yanjun Gao, PhD \corref{cor1}%
\fnref{fn1}}
\author[inst2]{Dmitriy Dligach, PhD}
\author[inst3]{Timothy Miller, PhD}
\author[inst1]{John Caskey, PhD}
\author[inst4]{Brihat Sharma, MS} 
\author[inst1]{Matthew M. Churpek MD, MPH, PhD; Majid Afshar, MD, MSCR }

\address[inst1]{ICU Data Science Lab, Department of Medicine, University of Wisconsin Madison}
\address[inst2]{Department of Computer Science, Loyola University Chicago}
\address[inst3]{Boston Children’s Hospital, Harvard University}
\address[inst4]{Department of Psychiatry and Behavioral Sciences, Rush University Medical Center}

\cortext[cor1]{Corresponding author}
\fntext[fn1]{Email author: ygao@medicine.wisc.edu}

            
            
            

\begin{abstract}
The meaningful use of electronic health records (EHR) continues to progress in the digital era with clinical decision support systems augmented by artificial intelligence. A priority in improving provider experience is to overcome information overload and reduce the cognitive burden so fewer medical errors and cognitive biases are introduced during patient care. One major type of medical error is diagnostic error due to systematic or predictable errors in judgement that rely on heuristics. The potential for clinical natural language processing (cNLP) to model diagnostic reasoning in humans with forward reasoning from data to diagnosis and potentially reduce cognitive burden and medical error has not been investigated. Existing tasks to advance the science in cNLP have largely focused on information extraction and named entity recognition through classification tasks. We introduce a novel suite of tasks coined as Diagnostic Reasoning Benchmarks, \textsc{Dr.Bench}, as a new benchmark for developing and evaluating cNLP models with clinical diagnostic reasoning ability. The suite includes six tasks from ten publicly available datasets addressing clinical text understanding, medical knowledge reasoning, and diagnosis generation. DR.BENCH is the first clinical suite of tasks designed to be a natural language generation framework to evaluate pre-trained language models for diagnostic reasoning. The goal of DR. BENCH is to advance the science in cNLP to support downstream applications in computerized diagnostic decision support and improve the efficiency and accuracy of healthcare providers during patient care. We fine-tune and evaluate the state-of-the-art generative models on DR.BENCH. Experiments show that with domain adaptation pre-training on medical knowledge, the model demonstrated opportunities for improvement when evaluated in DR. BENCH. We share DR. BENCH as a publicly available GitLab repository with a systematic approach to load and evaluate models for the cNLP community. We also discuss the carbon footprint produced during the experiments and encourage future work on DR.BENCH to report the carbon footprint.  
\end{abstract}



\begin{keyword}
Natural language processing \sep clinical diagnostic reasoning \sep clinical diagnostic decision support \sep clinical natural language processing benchmark  
\end{keyword}

\end{frontmatter}



\section{Introduction}\label{sec1}

Healthcare providers frequently update the care plan for patients through the electronic health records (EHR), which are designed to assist the workflow of clinical decision making via easy access and retrieval of the patient's medical data~\cite{fowler2014,brown2014}. However, EHRs also serve as a billing tool and unnecessary information is copied and pasted into the note contributing to note bloat~\cite{alpert2019, aronson2019}. 
Information and cognitive overload subsequently occur and contribute to missed diagnoses and medical errors~\cite{furlow2020, Hultman2019}. The National Academy of Medicine (NAM, formerly known as Institute of Medicine), showed that medical errors are the sixth leading cause for deaths~\cite{donaldson2000}, and diagnostic error is one of the more frequent types of medical errors~\cite{nas2015a}. Several recent studies discuss the possibility of reducing diagnostic errors using health information technologies to help to offload cognitive burden and biases~\cite{delvaux2020, AHRQ2019, nas2015b}.

In 2006, the first clinical natural language processing (cNLP) shared tasks were introduced in the Informatics for Integrating Biology and the Bedside (i2b2). Initial tasks were designed to apply NLP on EHR data for extraction and research purposes that demonstrate proof-of-concept and accurately apply NLP methods in the clinical domain. In accordance came the introduction of the first publicly available corpus of EHR notes from the Medical Information Mart for Intensive Care (MIMIC), which provided an increase of annotated datasets for cNLP tasks~\cite{johnson2016}. A scoping review of publicly available English language tasks identified 48 cNLP tasks based on EHR data between 2006 and 2021. Forty-seven percent were named entity recognition (NER) and information extraction (IE) tasks, which remain the predominant method in cNLP today~\cite{gao2022a}. Only a few tasks were intended for clinical applications, such as disease phenotyping and extracting risk factors for poor health outcomes but they remain in the realm of information extraction~\cite{gao2022a}. Although a small number of cNLP tasks were introduced in recent years to address medical knowledge representation and inference~\cite{romanov2018, yue2020}, a gap remains between cNLP models and applications that support clinical decision-making, in particular, to \textit{generate} diagnostic recommendations given the patients' information\cite{gao2022a, lederman2022}. A paradigm shift in cNLP is needed to connect the advanced NLP methods with a suite of tasks that could facilitate the development of newer models for clinical decision generation, ultimately clinical decision support tools that model human reasoning and synthesize data into real-time medical diagnosis to assist bedside care~\cite{lederman2022}.   


The \textbf{D}iagnostic \textbf{R}easoning \textbf{Bench}mark (\textit{DR.BENCH}), is intended to fill the gap and provide a new benchmark of clinical NLP tasks to facilitate model development and evaluation towards computerized clinical diagnostic decision support. The theoretical foundation of the proposed benchmark is \textit{Clinical Diagnostic Reasoning}, a critical and complex cognitive process defined in medical education that enables human physicians to conclude diagnosis and treatment plans with background medical knowledge and the evidence, which is documented in clinical notes~\cite{barrows1980, bowen2006, monteiro2013}. cNLP models that may accurately assess a patient's condition and potentially overcome the provider's cognitive bias when rapid decisions need to be made in a busy hospital setting (e.g.,``decisional shortcut'') is a promising direction forward.  

\begin{figure}[h]%
\centering
\caption{Mapping cognitive skills for clinical diagnostic reasoning to clinical natural language processing tasks }\label{map}
\includegraphics[width=0.9\textwidth]{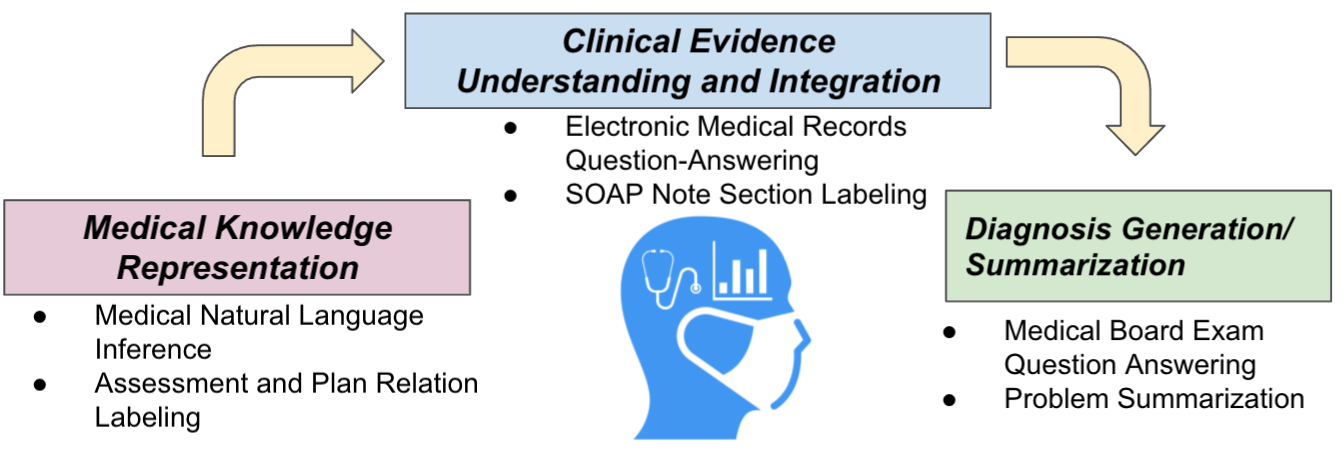}
\end{figure}

Different strategies for performing clinical diagnostic reasoning are proposed in the literature~\cite{barrows1980, Hammond2000, bowen2006, Pelaccia2011, monteiro2013}. The core elements are the ability to gather, understand and integrate clinical evidence, reason over the evidence using medical knowledge, and summarize relevant diagnoses. These cognitive skills are mapped to the following cNLP research areas: (1) medical knowledge representation, (2) clinical evidence understanding and integration, and (3) diagnosis generation and summarization. The three areas are also interdependent. Medical knowledge representation is fundamental to nonanalytic activities with a strong dependency on clinical experience that uses pattern recognition to formulate a diagnosis~\cite{rassinoux2008, blobel2013}. Clinical evidence represents a workup of diagnostic tests and gathering patient data as analytic reasoning alongside the experiential knowledge representation. Finally, the skill of understanding and integrating the clinical evidence with existing evidence-based medicine serves as the prerequisite to form decisions, i.e. diagnosis generation~\cite{bowen2006, Hutton2007}. Both knowledge representation and clinical experience are used simultaneously in an interactive fashion by clinicians and serve as the design for artificial intelligence systems to model. DR. BENCH incorporates cNLP tasks that cover all areas in the clinician's cognitive process (Figure~\ref{map}). The aim of DR. BENCH is to evaluate the progress of NLP models on clinical diagnostic reasoning and promote the design of models that may be applied as computerized diagnostic decision systems at the bedside.

\section{Methods}~\label{method}
 
\subsection{DR.BENCH: Diagnostic reasoning benchmark for clinical natural language tasks}
\begin{figure}[h]%
\centering
\caption{Introduction of DR.BENCH Framework with example input data and labels}\label{drbench}
\includegraphics[width=0.98\textwidth]{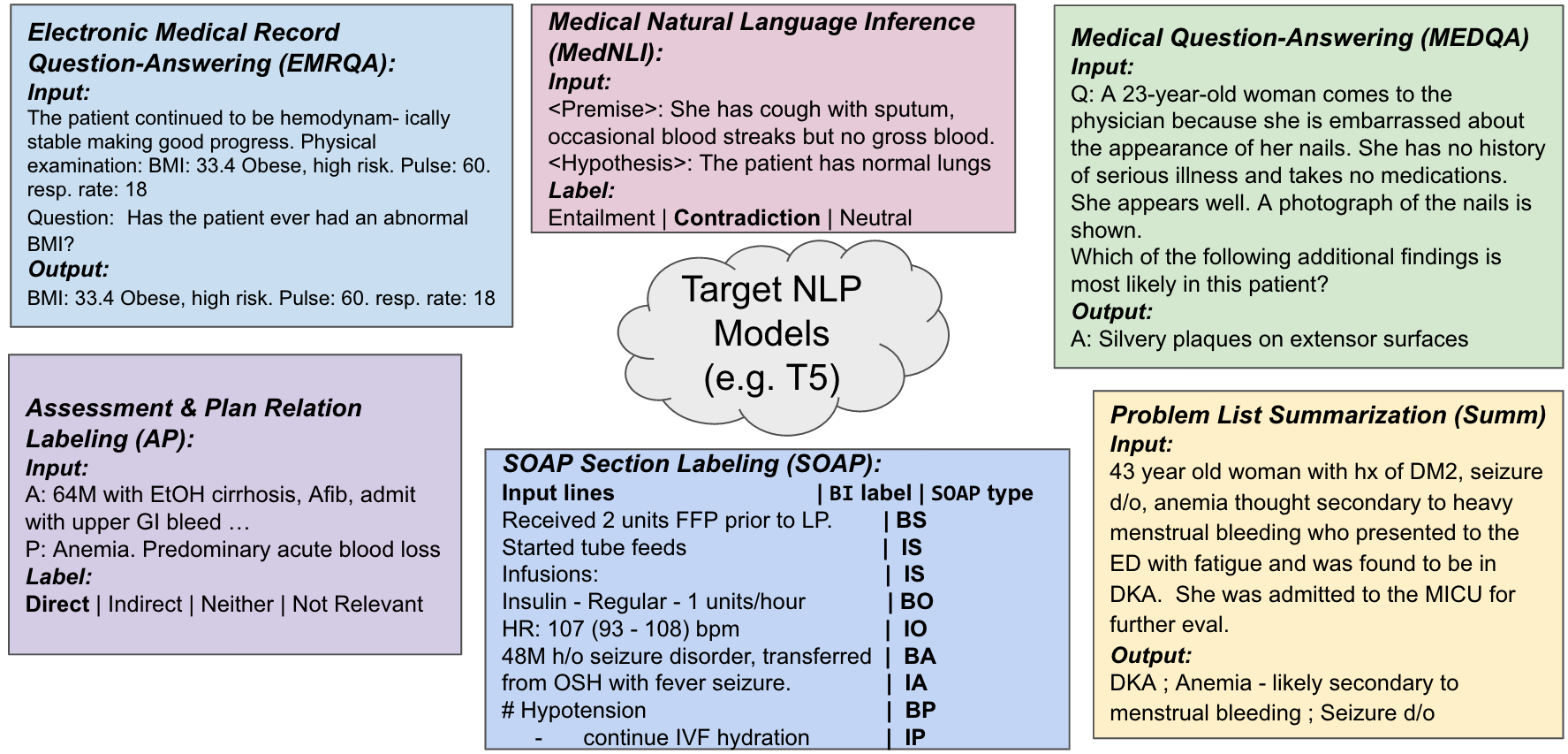}
\end{figure}

DR.BENCH was composed of six tasks from five existing publications with publicly available datasets~\cite{romanov2018,gao2022b,pampari2018,jin2021,gao2022c}.  
It was built upon our previous investigation of clinical NLP tasks that facilitated model development for clinical diagnostic reasoning. In our previous work, we designed a hierarchical annotation framework that followed the cognitive workflow of physicians reviewing the SOAP format daily progress note. We conducted and published three stages of annotation that corresponded to three clinical NLP tasks~\cite{gao2022b}: SOAP Section labeling (SOAP Labeling), Assessment and Plan Relation Labeling (AP), and Problem List Summarization (Summ). Each task addressed at least one aspect of the cognitive skills required for clinical diagnostic reasoning, presented in Figure~\ref{map}. In addition to the new tasks, we published a scoping review that examined 48 existing clinical NLP tasks that use public English EHR data, and identified the cNLP tasks that addressed clinical text understanding, medical knowledge representation and reasoning~\cite{gao2022a}.  Electronic medical records question-answering (EMRQA) and medical natural language inference (MedNLI) were incorporated into DR.BENCH. Finally, an additional task on medical board exam question answering (MedQA) was found and included to represent the prerequisite for conducting clinical practice and help further evaluate the qualification of a medical AI system~\cite{jin2021}. Figure~\ref{drbench} presents the example input and output for each task in DR.BENCH.  

The selection of the tasks covered a range of text units, beginning with sentence level and advancing to full-length daily care notes (e.g., daily progress notes). Most datasets (n=5) were sourced from MIMIC-III as the only fully deidentified and public EHR corpus of notes at the time of this publication. The majority of the tasks incorporated abstractive reasoning to test the medical knowledge of cNLP systems and move beyond the extraction of medical information directly from the corpus of text. One of the challenges of this benchmark was the tasks were conceived differently, as classification versus sequence generation. The goal was to unify these diverse task types as sequence generation, leveraging the power of recent large pre-trained generative models such as T5 (introduced in section~\ref{model})~\cite{raffel2020}. 
The following sections provide a detailed introduction to each task, including the task setup, data source and evaluation metric.

\subsubsection{TASK 1: MedNLI}  Natural language inference (NLI) is a task that one is given a ``Premise'', to determine its logical relation to a ``Hypothesis'': \textsc{Entailment}, \textsc{Neutral} or \textsc{Contradiction}~\cite{romanov2018}.  Figure~\ref{drbench} (top-centered) contains an example to represent the MedNLI task. The ``Premise'' contained ``cough with sputum, occasional blood streaks'' which indicated a problem in the respiratory system and \textsc{Contradicts} to the statement of ``Hypothesis'' that ``The patient has normal lungs''. To accurately predict the relations, models would need to generate precise semantic representation and then establish the inference between the meanings of the sentence pairs. Therefore, we categorized MedNLI as a task to assess the model's medical knowledge representation. Two board-certified radiologists provided the annotations and the results were reported in DR. BENCH as accuracy between the exact match on the labels and the generated text.      

\subsubsection{TASK 2: Assessment and plan relation labeling} 
The Assessment and Plan sections of the progress notes were the free-text fields where healthcare providers identified patients’ problems/diseases and treatment plans. The Assessment section summarized the patients’ active health problems or diseases from a single progress note for that day. The Plan section consisted of multiple subsections, each addressing a specific problem or diagnosis followed by a detailed treatment plan. In this task, each plan subsection was labeled with one of the four relations that indicated the nature of its association with the assessment: \textsc{Direct}, \textsc{Indirect}, \textsc{Neither}, \textsc{Not relevant}. Each label indicated if the disease/problem stated in each part of the plan subsection was a primary diagnosis or problem, a secondary problem, a problem that was not mentioned in the note, or not considered a diagnosis or problem~\cite{gao2022b}. 

Given the assessment and plan subsection as input, the task was to predict the four labels. Figure~\ref{drbench} (bottom-left) illustrates the task setup of Assessment and Plan relation labeling: the Plan subsection (\textsc{P}) mentioned ``anemia'', which was the main cause of ``EtOH cirrhosis'' and ``upper GI Bleed'' in the Assessment (A), and the label was \textsc{direct}. Similar to MedNLI, the model needed to generate a precise representation for Assessment and Plan subsections, then predicted the relation between them. Thus, this task was categorized as an evaluation of medical knowledge representation. This task was part of the National NLP Clinical Challenges (N2C2~\cite{n2c2}). To be consistent with N2C2, we reported Macro F1 as the evaluation metric for DR. BENCH.

\subsubsection{TASK 3: EmrQA} \label{sec:emrqa} EmrQA stands for electronic medical records question answering~\cite{pampari2018}. Given a clinical note and a question, the task was to extract a continuous text span from the clinical note as the answer. Previous work showed that EmrQA served as a resource for machine reading comprehension, an NLP task to identify, understand and integrate specific information from the input text~\cite{yue2020}. Figure~\ref{drbench} (top-left) included an EmrQA example where the question was to find abnormal values reported in the input clinical text. Therefore, EmrQA was categorized as a task to assess clinical evidence understanding and integration in DR. BENCH. EmrQA used expert annotations from five i2b2 challenges, including relations, medications, heart disease risk, smoking, and obesity. The developers of the task generated questions and answers using logical forms from the annotated templates. Results in DR.BENCH were reported as the accuracy in the exact match on the text span, and the resultant evaluation metric was the average accuracy over the five i2b2 datasets.    

\subsubsection{TASK 4: SOAP labeling} \label{sec:soap} The SOAP labeling task was to identify a note section into one of the following categories: (1) Subjective; (2) Objective; (3) Assessment; and (4) Plan. The SOAP note is a ubiquitous format in medical note writing that remains the foundation in medical education for organizing a clinical note \cite{weed1968}. Recent work showed that automated identification of SOAP sections helped physicians quickly locate specific information, especially the Assessment and Plan sections where diagnoses and treatment plans were mentioned~\cite{Edinger2018, eisman2022}. The task was to identify, at the line-level of the note, the correct SOAP section as well as demarcate the beginning of the section. Thus, each line of the note was labeled into eight different groups (the four sections of SOAP and the Beginning [B] or Inner [I] of that section). If the line in the progress note was a part of the beginning of a section, it was labeled as either BS, BO, BA, or BP; otherwise, Inner (I)S, IO, IA, or IP. To promote generalizability, we removed sub-section headers (i.e., ``Medications'', ``Past Medical History'', ``Physical Exam'', etc.) that were unique to the note type or hospital setting.  

Figure~\ref{drbench} (bottom-centered) presents nine lines of text with the B labels and SOAP labels in an example note. The task was designed to segment the notes into SOAP sections and predicted the topics of the sections, which required models to generate the accurate semantic representation of the current lines and understand their topic coherency with previous lines. Thus, this task evaluated the model's capacity in understanding and integrating clinical text. A sliding window of 5 previous lines was added to the input dataset while training a generative model. Two trained medical students provided the annotations, and the results in DR.BENCH were reported as the overall accuracy across the four sections and two positions.  

\subsubsection{TASK 5: MedQA}\label{medqa} MedQA was a large-scale question-answering dataset collected from a bank of practice medical board exam questions and answers~\cite{jin2021}. The corpus represented the question bank a medical trainee would read in the United States Medical Licensing Examination, which was a required step in assessing medical knowledge and reasoning for medical board certification. The task evaluated cNLP models in utilizing medical knowledge to answer the question. Given a question, the model predicted the correct answer from five answer options (A/B/C/D/E). An example of MedQA is presented in Figure~\ref{drbench} (top-right). Along with the previously published dataset was a collection of medical textbooks. We adapted the original MedQA task into two settings: \textit{open-book} a question and some relevant paragraphs from the textbook collection were given (an open-book simulation), and \textit{closed-book} where only the question was given (a closed-book simulation). In the open-book MedQA, we used BM25, an information retrieval model based on TF-IDF that returned a collection of documents for a query. BM25 was used to retrieve the top five paragraphs given a question~\cite{trotman2014}. For the closed-book MedQA, DR.BENCH only evaluated the model's internal knowledge representation to answer the question. Final results were reported as overall accuracy in generating the correct letter answer option.   

\subsubsection{TASK 6: Problem summarization} 
Given a progress note, the goal of the Problem Summarization task was to identify and generate the problems and diagnoses for the patient’s daily hospitalization. We provided two settings for this task: the first configuration took only the Assessment section in the progress note as input, because the Assessment section synthesized the evidence from the Subjective and Objective sections and contained information about the patient's current status. The Assessment setting was denoted as \textsc{Summ-Assmt} (a data example of \textsc{Summ-Assmt} is presented in the bottom-right of Figure~\ref{drbench}). In the second setting, all sections except the Plan section (because it contains the target problems/diagnoses) were provided as the input, denoting the task as \textsc{Summ-Note}. The progress note included free-text fields from Subjective and Assessment sections and semi-structured text from Objective sections such as lab results and vital signs, making \textsc{Summ-Note} the hardest task in DR.BENCH with the largest input of text. The data source and annotation were previously described~\cite{gao2022b}. ROUGE-L on the generated problems/diagnoses was the evaluation metric~\cite{lin2004}.  

\subsection{Baseline Experiments: Pre-trained models and domain adaptation pre-training} \label{model}

DR.BENCH was designed in a generative framework and a pre-trained seq2seq transformer, Google's Text-To-Text Transfer Transformer (T5)~\cite{raffel2020}, served as the baseline model across all tasks. T5 can handle numerous types of tasks through its flexible architecture and achieved state-of-the-art results on multiple language tasks~\cite{raffel2020}. Recently T5 has been used for clinical text generation.~\cite{gao2022c} T5 was trained on the Colossal Clean Crawled Corpus (C4), a text corpus comprised of 805 gigabytes of web data. Two T5 checkpoints were selected for experiments: T5-Base with 220 million parameters and T5-Large with 770 million parameters (T5-Base\textsc{-Vanilla} (T5-B-\textsc{Vanilla}) and T5-Large-\textsc{Vanilla} (T5-L-\textsc{Vanilla})).  

A study by Gururangan et al \citep{gururangan2020} demonstrated the performance gained from Domain Adaptation Pre-training (DAPT) from a second phase of pre-training on unlabeled data that were domain-specific. The multi-phase pre-training mechanism outperformed direct fine-tuning. Using similar methods, experiments in DR. BENCH included domain adaptation pre-training using T5 on two medical knowledge sources as in-domain corpora. The goal was to examine whether medical domain pre-training was useful. The two medical knowledge corpora were PubMed and Unified Medical Language System (UMLS). The selection of in-domain corpora represented the conventional methods of training language models for medical knowledge representation. In addition, we used EHR progress notes from MIMIC as another in-domain pre-training corpus that represented clinical experience. Clinical pre-trained language models (PLM) such as BioBERT,  ClinicalBERT, and SapBERT used PubMed, MIMIC, and UMLS, respectively, for pre-training ~\cite{lee2020biobert, alsentzer2019, liu2021sapbert}. The following models were designed for DR. BENCH and intended to establish baseline results.

\paragraph {Model 1a and 1b: Original (Vanilla) T5 and SciFive} All experiments began with the original, vanilla T5-B and T5-L models without any modifications. One of the first health domain T5 variants was called \textsc{SciFive}~\cite{phan2021}. \textsc{SciFive} was continuously trained on 32 million PubMed abstracts and full-text articles. In addition to Vanilla T5, we also examined \textsc{SciFive}-Base and \textsc{SciFive}-Large that corresponded with the T5-B and T5-L checkpoints.

\paragraph {Model 2: UMLS Concept definitions and medical textbook} The Unified Medical Language System (UMLS) was constructed and managed by the National Library of Medicine, and it is the largest curated knowledge source containing biomedical concepts and their relationships and definitions~\cite{Bodenreider2004}. We continually trained T5 on the data available in the 2022AA full UMLS release files (UMLS Metathesaurus and Semantic network) 
~\cite{Bodenreider2004}. The knowledge sources from the UMLS were organized into a Domain Adaptation Pre-training (DAPT) corpora in the following two ways: (1) extracted concept definitional sentences; and (2) extracted concept and relation triples. Prior studies used the UMLS knowledge source to improve sentence representation using definitional sentences from the UMLS word dictionary \cite{Tsukagoshi2021, guo2012}. In a similar fashion, we extracted all concept definitions as the DAPT corpora. The resultant dataset contained over 300,000 medical definitions that were appended to the medical textbook collections from MedQA (Section \ref{medqa}). The final size of the DAPT corpora was 515,000 training instances and referred to as \textsc{T5-Defs} for T5-B and T5-L experiments.

\begin{figure}[h!]%
\centering
\caption{ Model 3 of UMLS Concept Relation Paths (T5-B-\textsc{RelPaths} and T5-L--\textsc{RelPaths}). An example illustrating the training sample construction given concepts and relations from UMLS. The resulting training samples are used to continuously train T5. }\label{graph}
\includegraphics[width=1\textwidth]{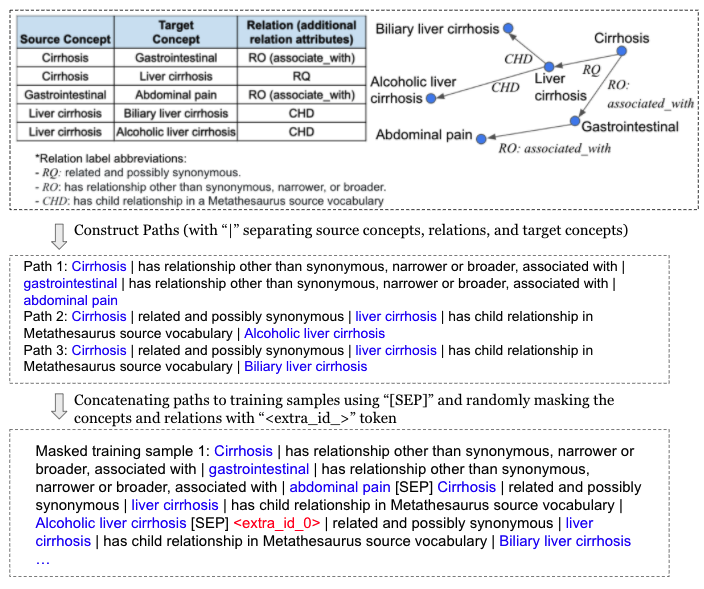}
\end{figure}

\paragraph{Model 3: UMLS concept relation paths} Medical concept relation information from the UMLS knowledge maps represent multi-hop relational chains that depict medical knowledge and reasoning. We attempted to learn these paths with continuous training on T5-B and T5-L using all medical concepts under the UMLS semantic type for ``Disease and Symptoms'' and their relation triples as $<concept_{source}, concept_{target}, relation>$. A directed graph was constructed with the nodes as the concepts and the edges as the relations. A graph traversal algorithm was executed to retrieve all the paths that consisted of two edges. The paths provided connections of concepts that were not direct neighbors but were linked through the intermediate nodes. Figure~\ref{graph} presents three example paths constructed from the source concept ``cirrhosis''. An example path was from ``cirrhosis'' to ``abdominal pain'', where ``cirrhosis'' was connected to ``gastrointestinal'', which had a direct edge to abdominal pain.   

The mean number of tokens in a given path was 47. For every source concept, we concatenated 10 paths into one training sample to avoid exceeding the 512 input token limit for T5. Each path was separated by the ``[SEP]'' token, indicating they were different paths. The final set of DAPT corpora contained approximately 582,000 training instances, roughly corresponding to 5.8 million paths and referred to as \textsc{T5-RelPaths}.  

\paragraph{Model 4: MIMIC progress notes} The daily progress notes from the MIMIC-III dataset were extracted and used to continually train T5-B and T5-L. Progress notes are clinical notes that document the patient's daily events and exam findings with the diagnoses and active problems followed by a treatment plan. The progress note is frequently formatted in the S-O-A-P format, where S (Subjective sections) and O (Objective sections) document the collected medical data and clinical evidence, and the Assessment (A) and Plan (P) sections contain diagnoses and treatment plans. To obtain high-quality progress notes for pre-training, we focused on provider notes in the SOAP format and extracted the ``Assessment and Plan'' sections that contained the diagnoses and related treatments. The subset of notes that were in the DR.BENCH test set was excluded during training. The final corpus contained 283 training examples and was designated as T5-\textsc{Ehr}.

\begin{figure}[h]%
\centering
\caption{Domain adaptation pre-training setup for DR.BENCH}\label{dapt}
\includegraphics[width=1\textwidth]{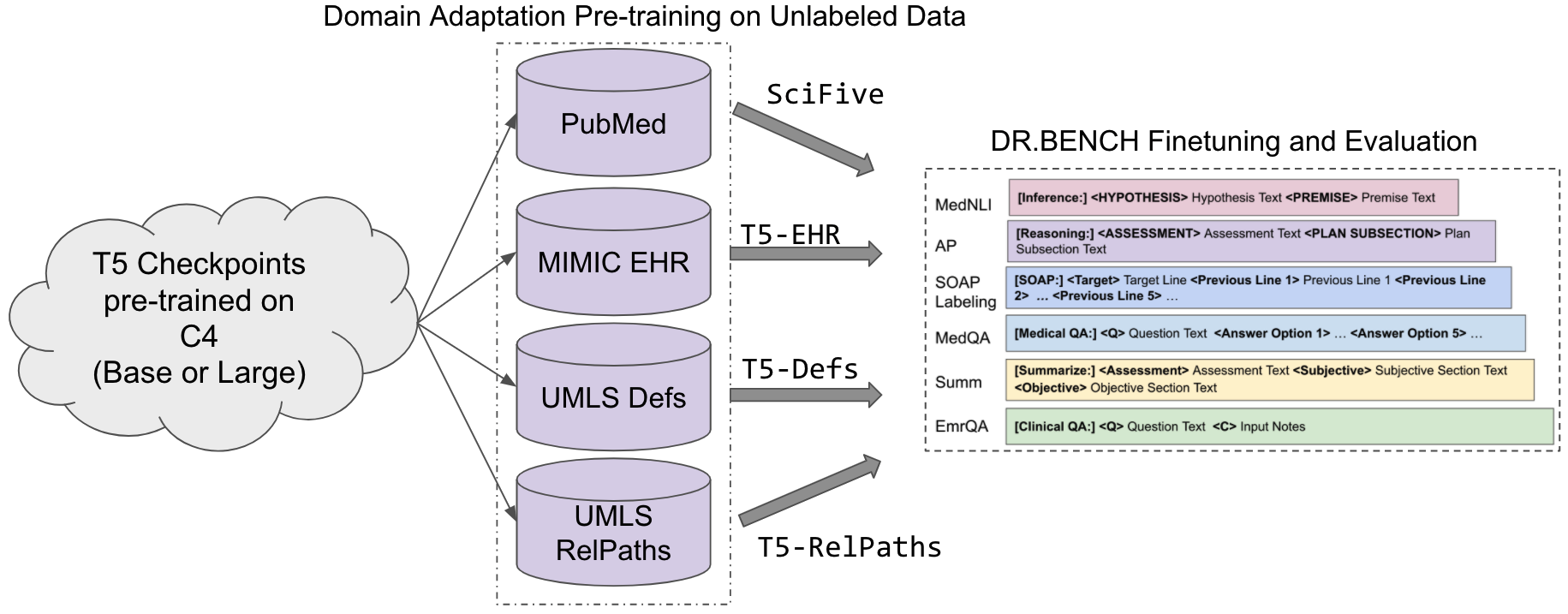}
\end{figure}

Figure~\ref{dapt} presents a workflow of experiments with the DAPT setup for DR.BENCH. The codebase of DR.BENCH was designed to support model fine-tuning and evaluation on all tasks in a single, consolidated framework. We continuously trained T5-B and T5-L with different domain-specific corpora. DR.BENCH took in the DAPT model checkpoints for further fine-tuning on the task training sets. 


\begin{table}[h]
\caption{T5 Domain Adaptation Pre-training (DAPT) Corpora Comparison} 
 \small 
 \begin{adjustbox}{width=1\textwidth}
    \begin{tabular}{l|l|l|l} \toprule
      Abbreviation   & Corpora Descriptions & Size & Masking Strategy   \\ \midrule 
       \textsc{Ehr} & MIMIC progress notes & 283K examples & Random concept masking \\ 
        \textsc{Defs} & UMLS concept definitions   & \multirow{2}{*}{515K examples} & \multirow{2}{*}{Random token masking} \\ 
        & and medical textbooks & & \\
        \textsc{RelPaths} & UMLS 2-hop relation paths   &  582K examples  & Random source/target   \\
              & for all ``Disease and Symptoms''  &(5.8M 2-hop paths) & concept masking and \\ 
              & concepts & & relation masking\\ 
        \textsc{SciFive} & PubMed abstracts and full-text & 32M abstracts & Random token masking\\
      \bottomrule
      \multicolumn{4}{l}{*We included T5-\textsc{Vanilla}, the original model trained on Colossal Clean Crawled Corpus (C4) } \\ 
      \multicolumn{4}{l}{ with 364M examples. We continuously trained it on the domain adaptation pre-training corpora } \\ 
      \multicolumn{4}{l}{ listed above.}
    \end{tabular}
     \end{adjustbox}
    \label{tab:dapt-corpora}
\end{table}

\subsection{Corpora comparison}
The characteristics, size, and masking strategy of the DAPT corpora are summarized in Table~\ref{tab:dapt-corpora}. All corpora were continually trained on top of the original, Vanilla T5-B and T5-L. Vanilla T5 was included in the experiments for comparison against the clinical T5 models. The selection of experiments covered a wide range of medical knowledge sources and text that were the major corpora for medical domain pre-training. The medical domains included physician-written clinical notes (MIMIC-III progress notes), medical vocabulary (UMLS), and scholarly papers in the biomedical domain (PubMed). C4 was the largest corpus used to pre-train T5 followed by the PubMed collections as the largest medical corpus for continual pre-training. 

\subsection{Previous results reported in literature}
The SOAP Section task, Assessment and Plan Relation task, and Problem List Summarization task were three new tasks designed for DR. BENCH without prior baseline results. For MedNLI, the benchmark performance had an accuracy of 86.57 by SciFive-Large~\cite{phan2021}. For EmrQA, existing work reported performance on subsets of i2b2 topics. For example, Yue et.al~\cite{yue2020} reported the highest scores of 25.68 and 86.94 on exact matching for Medications and Relations, respectively. Most recently, Li et. al \cite{li2022clinical} reported 30.2, 91.1, and 69.8 exact matching scores on Medications, Relations, and Heart Disease (Risk). For MedQA, knowledge graphs integrated with transformers provided an accuracy between 45.0 and 47.5~\cite{yasunaga2021, yasunaga2022}. However, MedQA was previously evaluated as a 4-way multiple-choice question-answering task. DR.BENCH provided the 5-way multiple-choice task that was in the original paper~\cite{jin2021}, and results for EmrQA used all five i2b2 topics.

\subsection{Experiments Setup}
\subsubsection{Input representation}
\begin{figure}[h]%
\centering
\caption{Input setting with task prefix (square brackets) and special tokens (brackets) indicating different parts of input text for DR.BENCH. Tasks are listed in the ascending order of average input length. }\label{input}
\includegraphics[width=0.9\textwidth]{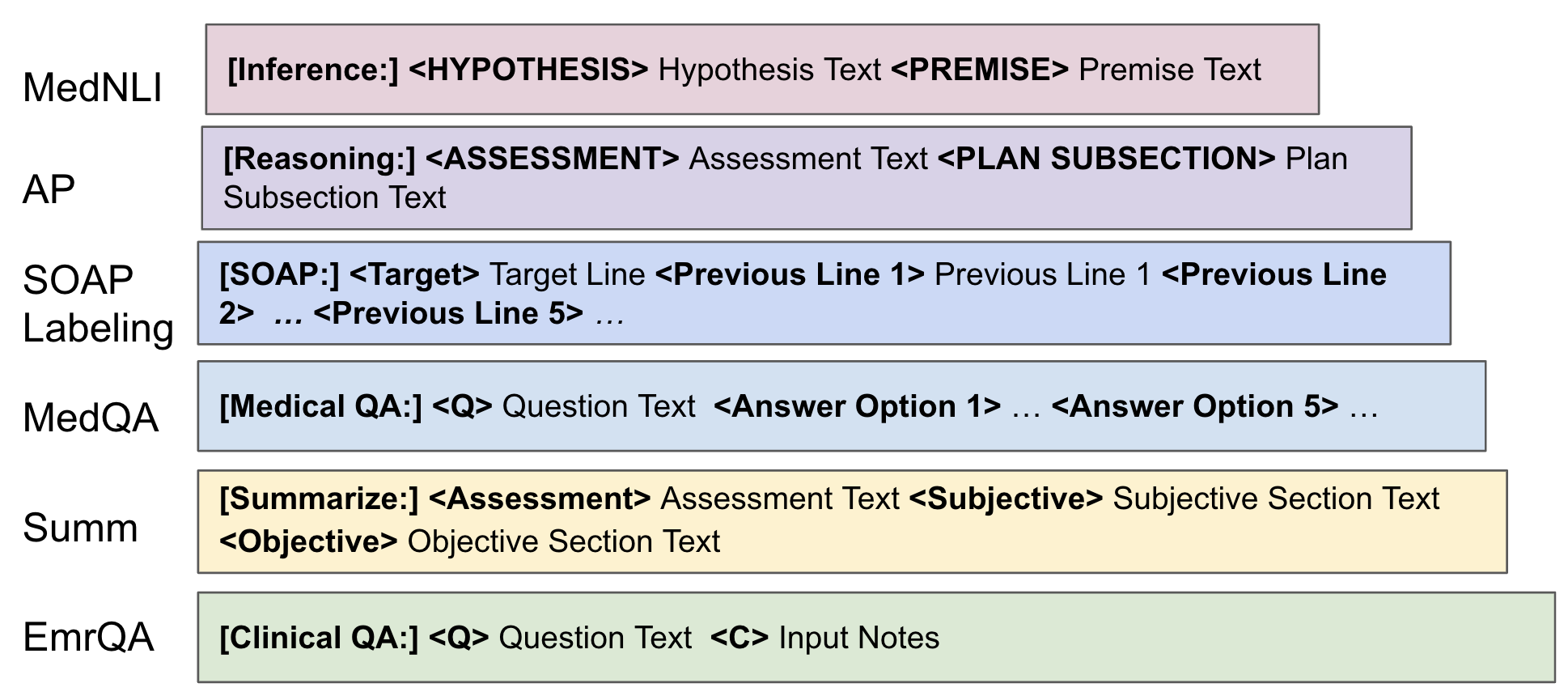}
\end{figure}

The input to T5 began with a short phrase (task prefix) to specify the task T5 would execute. All tasks had different parts of the text as input; therefore, we applied custom tokens to indicate the text source, as shown in Figure~\ref{input}. Some tasks incorporated the entire clinical note so the T5 tokenizer truncated the text when the 512 token limit was met. For the Problem Summarization task, the Assessment section was prioritized in the token order because it contained the main problems and symptoms, followed by the Subjective and Objective sections.  

\subsubsection{Continous Pre-training}
T5 used a random token masking policy to perform masked language modeling with random replacement of text spans using the special tag ``$<extra\_id\_n>$". For T5-\textsc{Defs}, we adapted a concept masking strategy introduced in~\cite{gao2022c}. We first applied QuickUMLS, a medical concept extractor based on UMLS~\cite{soldaini2016} to extract the concepts, and randomly masked 15\% of the concepts and T5 recovered the masked concepts during DAPT. For T5-\textsc{RelPaths}, we applied random path masking that randomly replaced the source concept, target concept, or the relation with the special tag. The goal of the masking strategy was to have T5 learn the relation-conditioned concept information. Finally, we concatenated the \textsc{Defs} corpora with the \textsc{RelPaths} corpora and continuously trained T5-Large (T5-Large-\textsc{RelPaths}+Defs). 

\subsubsection{Hyper-parameters searching}
Hyperparameter tuning was conducted mainly on the learning rate. DR. BENCH included a learning rate grid-search for all models on all tasks. The learning rate was constrained between 1e-3 to 1e-6 as previously described ~\cite{raffel2020}. We committed to the learning rate that achieved the best validation performance. All tasks were set with early stopping to prevent over-fitting. The input length to T5 was set to 512 tokens. Depending on the length of the input and the memory usage, the batch size varied between 4 and 32.  

\begin{table}[h]
\caption{Hyperparameters for fine-tuning T5 on DR.BENCH} 
 \small 
    \centering
    \begin{tabular}{c|c} \toprule
      Hyper-parameter   & Setting   \\ \midrule 
      Optimizer   &  AdamW\\ 
      Epoch & 20 (with early stopping) \\ 
      Learning rate & 1e-3, 1e-4, 1e-5, 1e-6 \\ 
      Batch size & 4, 8, 16, 32 \\ 
     Encoder max length & 512 \\ 
     Decoder max length & 64 \\ 
     Beam size & 10 \\ 
     Length penalty & 1 \\ 
     no repeat ngram size & 2 \\ 
      \bottomrule
    \end{tabular}
    \label{tab:param_ft}
\end{table}

\subsubsection{Evaluation metrics}
Resampling techniques with 1000 bootstrap samples were used to estimate the 95\% confidence intervals (CI) for all evaluation metrics. Experiment results were reported in Table~\ref{emrqa_soap} for clinical text understanding, Table~\ref{AP_MEDNLI} for medical knowledge representation and reasoning, and Table~\ref{medqa_results} and Table~\ref{summ_results} for diagnosis generation and summarization.

Error analysis was on the EmrQA and Summarization tasks. These two tasks were considered difficult tasks because of their long document structure and inconsistent formatting with sentence fragments and embedded structured data. EmrQA was collected over five years of i2b2 Challenges that addressed different clinical uses: relations, health risk factors, medications, smoking and obesity. The complexity varied for each subtask so separate analyses were performed across each subtask. For Summarization, we identified errors across examples for extractive and abstractive summarization.

\subsubsection{Computing infrastructure}~\label{infrastructure}
All experiments were executed in the Google Cloud Computing (GCP) platform, with a Linux-based Virtual Machine (VM) instance and 100GB Solid State Drive (SSD). We trained and fine-tuned all models on 2 to 4 Nvidia Tesla A100 GPUs with 40 GB GPU memory, depending on the cloud GPU availability. As part of the experiment results, we reported the GPU cost and carbon footprint of the experiments on GCP.  


\section{Results}

\begin{table}[]
\small 
    \begin{center}
    \caption{Characteristics (note types and input units) and statistics (input lengths, size of train/val/test set) of tasks in DR.BENCH. We report average numbers of tokens in the input as input length. Note that for Summ, we report the number of diagnoses as the size of train/val/test set. }
    \begin{adjustbox}{width=1.0\textwidth}
    \begin{tabular}{l|l|l|l|l|l|l} 
    \toprule
      Task  & Note type & Input unit & Mean Input Length & Train & Val & Test \\ \midrule 
       EmrQA  & Discharge summaries & Notes and questions & 1093.10 & 42607 & 5246 & 5346   \\ 
        SOAP Label & Progress notes & Lines of text & 25.82 & 106126 & 12957 & 15006 \\ 
        AP & Progress notes  & Paragraph pairs & 76.97 & 4633 & 597 & 667 \\
        MedNLI & Past medical history & Sentence pairs & 24.80 & 11232 & 1395 & 1422 \\
        MedQA & Medical licensing  & Paragraph and & 508.24 & 10178 & 1273 & 1274\\ 
         & exam questions & questions & & &  \\ 
        Summ & Progress notes & Notes & 423.88 & 2138 & 304 & 341\\ 
    \bottomrule
    \end{tabular}
    
    \end{adjustbox}
    \end{center}
    \label{stat}
\end{table}

\begin{table}[]
    \begin{center}
    \caption{Performance of T5 and its variants on tasks addressing medical knowledge representation. F1: Macro F1 score; Acc.: Accuracy; 95\% CI: Ninety-five percent confidence interval on bootstrapping samples.}\label{AP_MEDNLI}
    \small 
    \begin{tabular}{ll|l|l|l|l} 
    \toprule
       \multicolumn{2}{c|}{\multirow{2}{*}{Model}}  & \multicolumn{2}{c|}{AP} & \multicolumn{2}{c}{MedNLI}  \\ 
       &  & F1 & 95\% CI & Acc. & 95\%  CI 
       \\ \midrule 
       \multirow{4}{*}{T5-B} & \textsc{Vanilla} & 73.31 & 71.34-77.65 & 79.75 & 78.62-82.70 \\
       & \textsc{Ehr} & 76.52 &74.01-79.33 & 80.02 & 79.98-82.12\\ 
       & \textsc{RelPaths} & 74.81 & 71.76-78.18 & 80.66 & 77.50-81.65\\
       & \textsc{Defs} & 72.22 & 69.46-76.26 & 80.73 & 78.69-82.77 \\
       & \textsc{SciFive} & 76.76 & 74.81-80.92 & 82.84 & 80.87-84.74 \\
       \midrule 
       \multirow{4}{*}{T5-L} & \textsc{Vanilla} & \textbf{77.96} & 75.38-81.60 & 84.04 & 82.14-85.86 \\
       & \textsc{Ehr} & 80.09  & 79.32-83.23 & 83.33 & 82.29-86.03 \\ 
       & \textsc{RelPaths} & 75.14 & 71.88-78.21 & 83.68 & 81.79-85.58 \\
       & \textsc{Defs} & 77.51 & 74.31-80.54 & 83.76 & 82.00-85.79 \\
       & \textsc{RelPaths}+Defs & 77.66 & 75.98-81.96 & 84.25 & 82.35-86.08 \\
       & \textsc{SciFive} & 76.76 & 75.25-81.20 & \textbf{84.88} & 82.98-86.64 \\
    \bottomrule
    \end{tabular}
    \footnotetext[1]{F1: Macro F1 score; Acc.: Accuracy; 95\% CI: Ninety-five percent confidence interval on bootstrapping samples.}
    \end{center}
    
\end{table}

\begin{table}[]
    \begin{center}
    \caption{Performance of T5 and its variants on tasks addressing clinical evidence understanding and integration} \label{emrqa_soap}
    \begin{minipage}{\textwidth}
    \begin{tabular}{ll|l|l|l|l} 
    \toprule
      \multicolumn{2}{c|}{\multirow{2}{*}{Model}}   & \multicolumn{2}{c|}{EmrQA} & \multicolumn{2}{c}{SOAP Labeling}  \\ 
      &  & Acc. & 95\% CI & Acc. & 95\%  CI 
       \\ \midrule
       \multirow{4}{*}{T5-B} & \textsc{Vanilla} & 33.40 & 29.27-37.61 & \textbf{60.12} & 59.33-60.90 \\
       & \textsc{Ehr} & 35.89 & 35.50-38.23 & 57.89 & 56.98-59.21 \\ 
        & \textsc{RelPaths} & 34.05 & 29.97-38.57 & 58.85 & 58.06-59.63 \\
       & \textsc{Defs} & 34.57 & 30.16-39.03  & 58.59 & 57.81-59.40\\
        & \textsc{SciFive} & 37.28 & 32.84-42.11 & 57.74 & 56.95-58.53 \\
    \midrule 
       \multirow{4}{*}{T5-L} & \textsc{Vanilla} & 38.05 & 33.56-42.58 & 55.57 & 54.78-56.35 \\
       & \textsc{Ehr} &36.23 & 34.95-38.66 & 54.22 & 53.14-56.73 \\
       & \textsc{RelPaths} & 37.25 & 32.76-41.78 & 59.06 & 58.29-59.83 \\
       & \textsc{Defs} & 38.28 & 33.72-42.83 & 60.06 & 59.27-60.84\\
       & \textsc{RelPaths}+Defs & \textbf{39.20} & 34.63-43.78 & 58.54 & 57.75-59.33 \\
       & \textsc{SciFive} & 38.23  & 33.69-42.79 & 59.53 & 58.76-60.33 \\
    \bottomrule
    \end{tabular}
    \end{minipage}
    \end{center}
\end{table}

\begin{table}[ht!]
\small 
    \begin{center}
    \caption{Performance of T5 and its variants on medical board exam question-answering tasks. We report two settings of the task: MedQA-Open where the top 5 paragraphs relevant to the questions are provided as part of the input; MedQA-Closed where no additional information is given besides the questions. }\label{medqa_results}
 
    \begin{tabular}{ll|l|l|l|l} 
    \toprule
       \multicolumn{2}{c|}{\multirow{2}{*}{Model}}   & \multicolumn{2}{|c|}{MedQA-Open} & \multicolumn{2}{|c}{MedQA-Closed}  \\ 
       &  & Acc. & 95\% CI & Acc. & 95\%  CI 
       \\ \midrule 
       \multirow{4}{*}{T5-B} & \textsc{Vanilla} & 22.55 & 20.01-25.69 & 22.07 & 20.44-25.07  \\
       & \textsc{Ehr} & 21.23 & 19.32-25.66 & 22.69 & 21.03-24.21 \\ 
       & \textsc{RelPaths} & \textbf{24.59} & 22.31-27.02 & \textbf{23.02} & 20.74-25.29 \\
       & \textsc{Defs}  & 20.35 & 18.22-22.62  & 20.97 & 18.77-23.17  \\
       & \textsc{SciFive} & 22.78 & 20.50-25.14  & 22.62 & 20.35-24.90 \\
      \midrule 
       \multirow{4}{*}{T5-L} & \textsc{Vanilla} & 20.97 & 18.77-23.25 & 19.32 & 17.20-21.52  \\
       & \textsc{Ehr} & 23.33 & 19.68-24.69 & 19.64 & 17.44-21.52 \\
       & \textsc{RelPaths} & 24.35 & 22.07-26.79 & 20.03 & 18.34-22.14  \\
       & \textsc{Defs} & 22.70 & 20.42-25.06 & 20.27 & 18.07-22.55 \\
       & \textsc{RelPaths}+Defs & 21.60 & 19.40-23.96 & 21.21 & 18.93-23.49  \\
       & \textsc{SciFive} & 21.37  & 19.09-23.64 & 20.47 & 18.46-23.02 \\
    \bottomrule
    \end{tabular}
    \end{center}
\end{table}

DR.BENCH was fully automated and standardized for any generative model via a modular script that handled loading, fine-tuning, and evaluation across all six tasks with an automated output of the evaluation metrics and their 95\% CIs. The characteristics of each task, including the total sample size across the training, validation, and test datasets are described in Table~\ref{stat}. For instance, the SOAP Note Labelling task had a total of 106,126 labels in the training data extracted from 603 progress notes, 12,957 validation data extracted from 75 progress notes, and 15,006 testing data extracted from 82 progress notes.

\begin{table}[ht!]
\small 
    \begin{center}
    \caption{Performance of T5 and its variants on problem summarization task: \textsc{Summ-Assmt} denotes the task that takes the Assessment section as input; \textsc{Summ-Note} denotes the task that takes the entire note as input (except Plan sections). We report ROUGE-L (RL) for Summarization task. }\label{summ_results}

    \begin{tabular}{ll|l|l|l|l} 
    \toprule
       & Model & \multicolumn{2}{c|}{Summ-Assmt} &  \multicolumn{2}{c}{Summ-Note} \\ 
       &  &  RL& 95\% CI & RL & 95\% CI 
       \\ \midrule 
       \multirow{4}{*}{T5-B} & \textsc{Vanilla}   & 14.08 & 11.91-16.25  & 4.71 & 3.43-5.99  \\
       & \textsc{Ehr} & \textbf{18.72} & 16.36-19.67 & 4.69 & 3.45-5.88 \\ 
       & \textsc{RelPaths} & 14.15& 12.29-16.01 & 4.00 & 2.75-5.26\\
       & \textsc{Defs}   & 17.33 &	14.12-20.53 & 5.38 & 3.39-7.38 \\
       & \textsc{SciFive} & 11.13 &	9.10-13.16 & 2.14 & 1.41-2.86\\
      \midrule 
       \multirow{4}{*}{T5-L} & \textsc{Vanilla} & 12.55	& 11.02-14.07 & 3.64 & 2.50-4.77 \\
    & \textsc{Ehr} & 15.79 & 12.55-19.03 & \textbf{7.60} & 5.31-9.89 \\ 
       & \textsc{RelPaths} & 14.43 & 11.65-17.21 & 3.03 & 2.12-3.94 \\
       & \textsc{Defs} & 12.84 & 10.06-15.63 & 4.68 & 2.53-6.83 \\
       & \textsc{RelPaths}+Defs & 12.99 & 10.98-15.00 & 5.66 & 3.79-7.54 \\
       & \textsc{SciFive} & 10.68 &	8.89-12.47 & 3.24 & 2.14-4.35 \\
    \bottomrule
    \end{tabular}
    \end{center}
\end{table}

\begin{figure}[ht!]
    \centering
    \caption{Point plot with confidence intervals from four models based on T5-Large.}
    \includegraphics[width=0.85\textwidth]{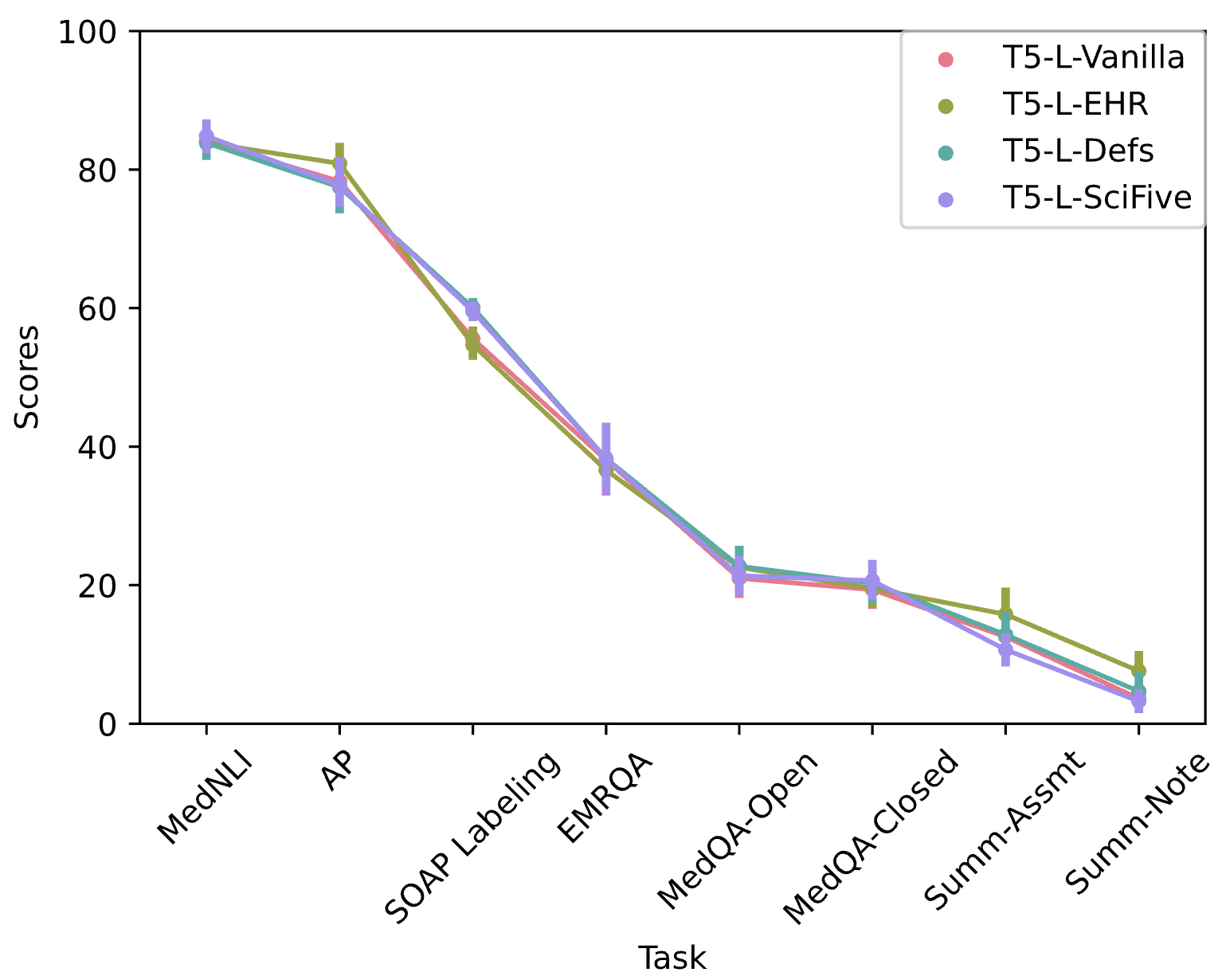}
    \label{fig:CI_figure}
\end{figure}

The models achieved the best performance on MedNLI with an accuracy range between 79.75\% and 84.88\% (Table~\ref{AP_MEDNLI}). Problem summarization (\textsc{Summ-Note}), which was intended as the most challenging task, had the lowest performance across all models, with Rouge-L scores between 2.14\% and 5.66\% (Table~\ref{summ_results}).  

A performance gain was observed when the base model changed from T5-B to T5-L, particularly on MedNLI, AP, EmrQA, and SOAP Labeling tasks. T5-L-\textsc{Vanilla} models demonstrated the largest gain over T5-B-\textsc{Vanilla} models on AP and MedNLI tasks, with +4.65 gain on AP macro F1 score and +4.29 gain on MedNLI accuracy.

T5-L-\textsc{Vanilla} achieved comparable performance on most tasks to models that were continually trained on medical data such as the T5-L DAPT variants. The exception was for the SOAP Labeling task where the continually trained T5-L models achieved nearly 60\% accuracy, approximately five points higher than \textsc{Vanilla}-T5 (Table 3). On most tasks, the 95\% CI of the \textsc{SciFive} variants overlapped with the three UMLS-based T5 variants. One exception was for \textsc{Summ-Assmt}, where T5-B-\textsc{Ehr} achieved a ROUGE score of 18.72, outperforming all models and 8.04 points above the best \textsc{SciFive} variant.       

To demonstrate the performance differences across models and tasks, Figure~\ref{fig:CI_figure} presents the 95\% CI of a subset of T5-Large system performance. Although the evaluation metrics were different across tasks, the range of scores was on the same scale between 0 and 100; therefore, we plotted the tasks together to further visualize the overlapping scores across models and between tasks. No single model provided marginal performance gain. T5-L trained on \textsc{Ehr} showed small gains over the Vanilla T5-L on the Summarization task.     

A total of 27 experiments were performed and resulted in Tables 2-4. The total toal cost was 5,300 USD with approximately 1,128 total GPU working hours. The MedNLI task had the shortest average input length and a batch size of 8 took 7GB of GPU memory to fine-tune T5-B, but the T5-L required 17GB of GPU memory. For long document tasks such as \textsc{Summ-Note}, the batch size was 4 and took 12GB GPU memory for the T5-B and 30GB for the T5-L. All experiments on the GCP produced 106 kg CO$_2$.  

\section{Error Analysis}

\begin{figure}
    \centering
    \caption{Accuracy of T5-L-\textsc{Ehr} and T5-L-\textsc{RelPaths+Defs} models on the five i2b2 topics in EmrQA. Test set size of each topic is included and upper bound error bars for each accuracy metric ($n$). }
    \includegraphics[width=0.9\textwidth]{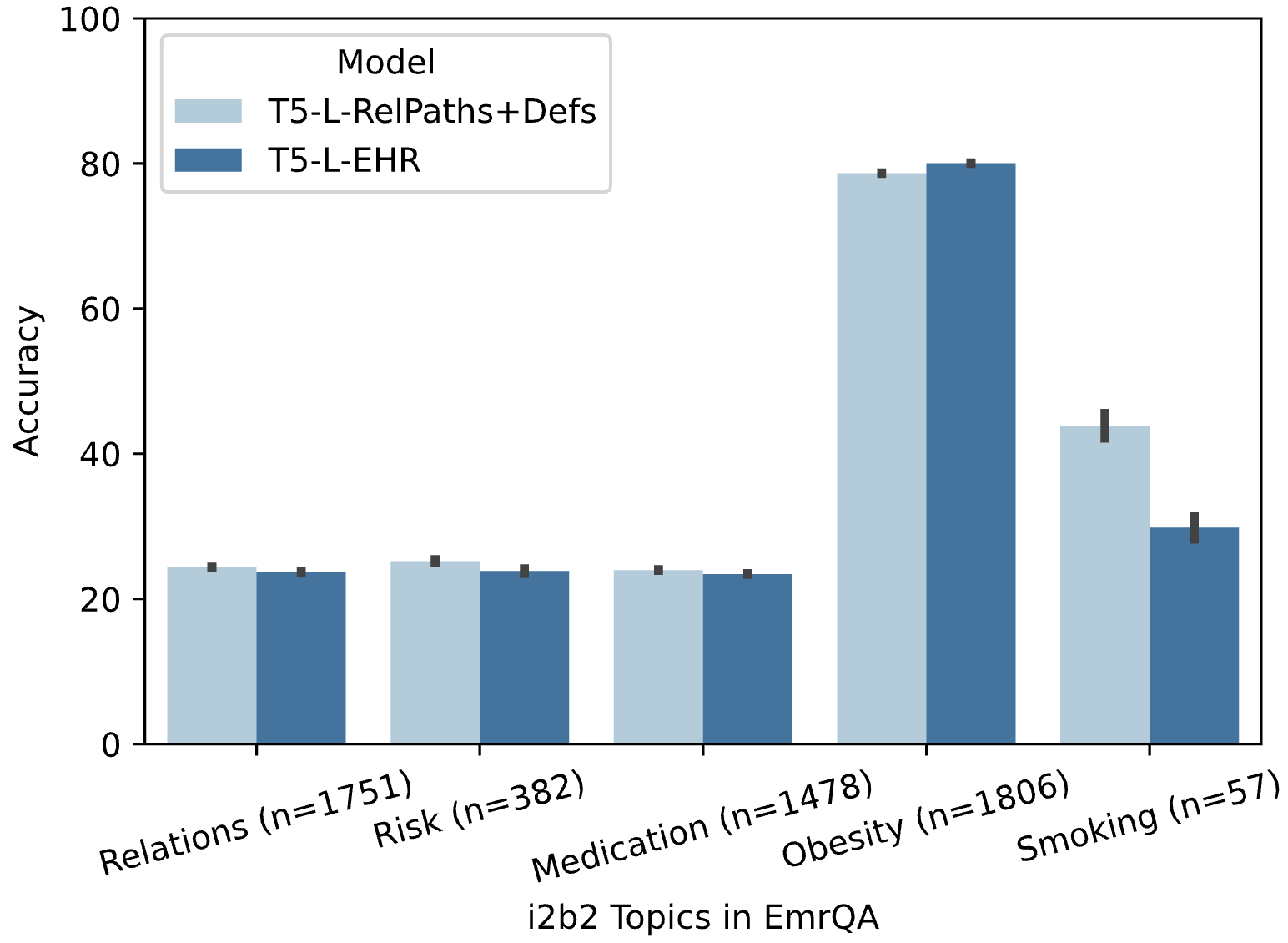}
    
    \label{fig:emrqa_error}
\end{figure}

\begin{figure}[ht!]
    \centering
    \caption{Input assessment, subjective sections and objective sections, and ground truth summary from an example progress note. There are 7 diagnoses in the ground truth summary, concatenated by semicolons. Texts highlighted in blue color are the diagnoses that are extractive. Model output on this input note is presented in the next figure (Figure~\ref{fig:outputerror}). }
    \includegraphics[width=1\textwidth]{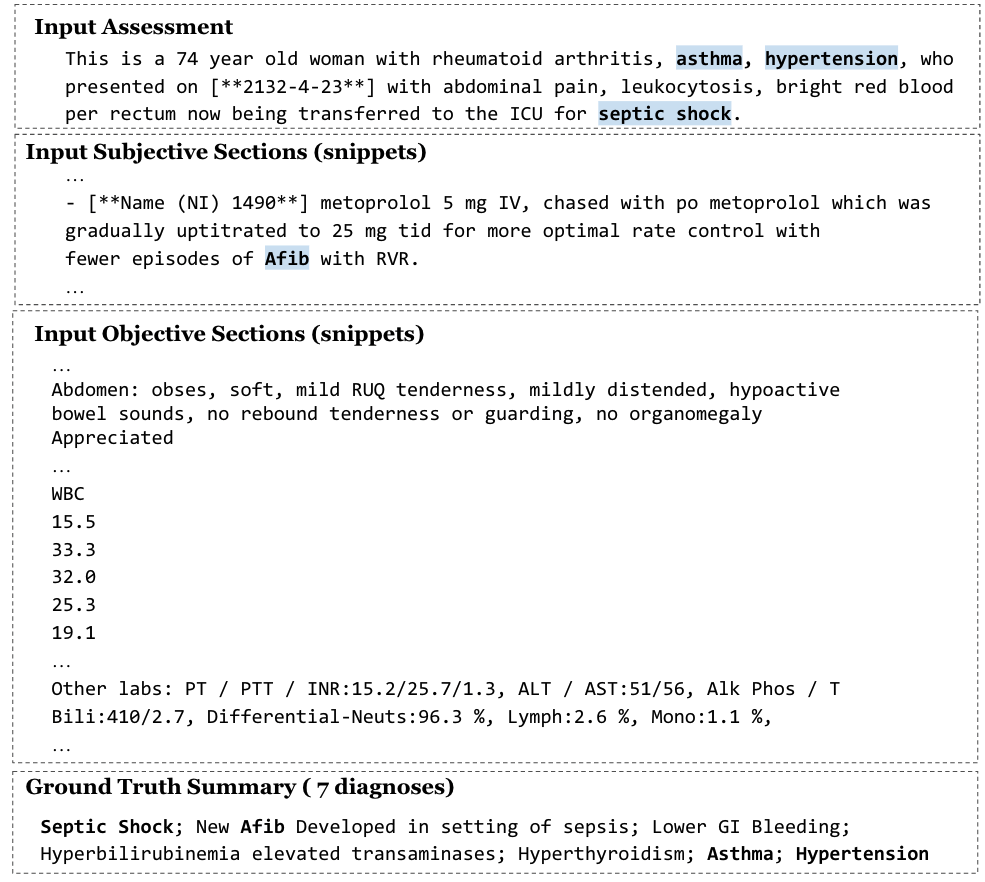}
    \label{fig:inputexample}
\end{figure}

The EmrQA task was to extract text spans from an input note to answer a related medical question. Figure~\ref{fig:emrqa_error} illustrates the differing performance of T5-B and T5-L across the five i2b2 topics in EmrQA. Most questions in the i2b2 Obesity Challenge were YES-NO questions (e.g., ``Does GERD exist'', ``Does the patient have any comorbidities associated with Obesity') that reflected a simpler task with a higher accuracy score than the other topics. Relations, Risk, and Medications were harder topics and answering questions in these three sets required understanding the content of questions and input notes, as well as locating the key text spans (e.g., ``What lab results does he have that are pertinent to recurrence of the tumor diagnosis'' (topic: Relation, answer: \textit{ct scan})).

Error analysis on the Summarization task was performed because it was perceived as the most challenging task with the lowest performance scores. Figure \ref{fig:inputexample} shows the input text starting with the Assessment followed by Subjective and Objective sections to meet the token limitation of a long document. The ground truth summary contained seven diagnoses, separated by semicolons. The blue highlighted diagnoses were extractive summarization diagnoses because they were explicitly mentioned in the input text. For the other diagnoses that had no explicit mention, they required \textit{abstractive summarization} with complex reasoning to conclude from the input. For example, ``Lower GI Bleeding'' could be induced from ``abdominal pain'' with ``bright red blood per rectum'' mentioned in the input assessment. Some diagnoses were harder to abstract. For example, ``Hyperbilirubinemia elevated transaminases'' is a form of liver dysfunction that requires information from the laboratory data, which was captured in the Objective section. To generate the diagnosis of ``Hyperbilirubinemia elevated transaminases'', a model would need to learn the relationship between the abnormal liver tests and their description as problems.

\begin{figure}[ht!]
    \centering
     \caption{Six sets of models output under two input settings on the given progress note presented in Figure~\ref{fig:inputexample}: Assessment input only (\textsc{Summ-Assmt}), and Assessment, Subjective sections, and Objective sections (\textsc{Summ-Note}). We use \textcolor{blue}{blue-font text} to highlight the correctly predicted diagnoses, and \textcolor{red}{red-font text} to highlight the words that are generated instead of extracted. }
    \includegraphics[width=1\textwidth]{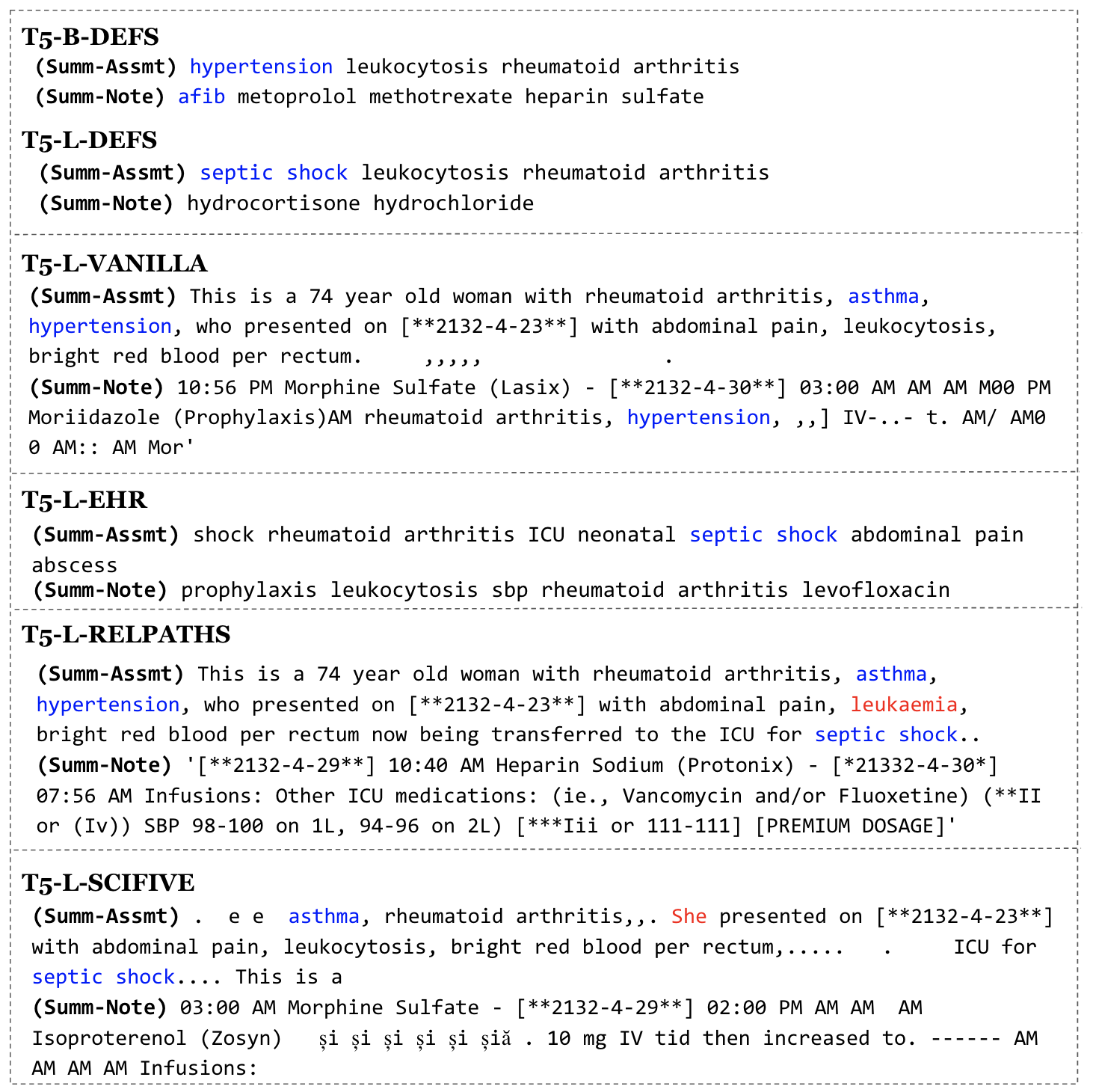}
    \label{fig:outputerror}
\end{figure}

Figure~\ref{fig:outputerror} presents six system outputs, and each contains two generated summaries: summary given the Assessment section as input only (``\textsc{Summ-Assmt}''), and summary given the input Assessment, Subjective and Objective sections (``\textsc{Summ-Note}''). The analyses contained two comparisons: 1) T5-B and T5-L models trained on the same corpora, and we compared T5-B-\textsc{Defs} with T5-L-\textsc{Defs}; 2) the same T5 checkpoint on different domain adaptation pre-training corpora, and we compared T5-L on \textsc{Vanilla}, \textsc{RelPaths}, \textsc{SciFive}, and \textsc{Ehr}. 

Across all examples, the quality of summaries was higher when the input was Assessment only (\textsc{Summ-Assmt}). T5-B-\textsc{Defs} predicted ``hypertension'' and T5-L-\textsc{Defs} predicted ``septic shock'' correctly. T5-L-\textsc{Vanilla}, T5-L-\textsc{RelPaths}, and T5-L-\textsc{SciFive} copied most of the text from input assessment for \textsc{Summ-Assmt} setting. For the \textsc{Summ-Note} setting, the output was mainly treatments and medications copied from the Objective section. T5-\textsc{L-Defs} and T5-L-\textsc{EHR} did better with extracted diagnoses with shorter text output. None of the models were able to generate abstractive diagnoses. Few ``re-writing'' behaviours were observed: T5-L-\textsc{RelPaths} incorrectly rewrote ``leukocytosis'' to ``leukaemia"; and T5-L-\textsc{SciFive} incorrectly rewrote the relative clause ``who presented on ...'' in Assessment to a sentence with a pronoun ``She''.

\section{Discussion}\label{discussion}
DR.BENCH is the first clinical NLP benchmark for clinical diagnostic reasoning proposed as a unified natural language generation framework and composed of six tasks across ten datasets. The suite of tasks covered discharge summary and progress note types, and input units were varied from shorter sentence-level to longer document-level. We evaluated a state-of-the-art generative model with T5 using the 220M parameters (T5-B) checkpoint and 770M parameters checkpoint (T5-L). We also compared medical domain-specific T5 variants that were continually trained on PubMed, EHR notes, and UMLS. The experiment results demonstrated tasks addressing medical knowledge representation and reasoning achieved the highest performance, and the group of tasks representing more diagnosis generation and problem summarization achieved the lowest, indicating the complexity was highest in summarizing problems/diagnoses. Error analysis highlighted examples with models that were unable to perform abstractive summarization and relied largely on extractive summarization. We showed T5-L-\textsc{Vanilla} was a difficult baseline to outperform despite our attempts at domain adaptation and continual training on medical knowledge databases. 


T5 achieved good performance on tasks addressing medical knowledge representation (AP and MedNLI), but the tasks of long clinical document understanding (EmrQA) and diagnosis generation and summarization (MedQA, Summ) remained challenging. Clinical note summarization (\textsc{Summ-Note}) takes both free-text fields and structured data as input and is considered to be the most complex task in DR.BENCH. Solving this task requires not only information integration and abstraction over long document, but also multi-modal understanding, going beyond T5's ability. All models had low accuracy between 20\% to 24\% on MedQA, which is the task that a medical student is required to achieve in the pathway to medical certification. The low accuracy on Open-book MedQA may partially be attributable to the errors propagated from the information retrieval algorithm. Nevertheless, the closed-book setting without information retrieval was also low. Continually training T5 on textbook and academic articles did not improve MedQA performance and suggests T5 is not learning or memorizing through ``reading" the textbook, representing similar conclusions made in general domain question answering~\cite{ciosici2021}. The performance on MedQA and Summ was consistently worse than the performance on AP and MedNLI. This may suggest that the models have the ability to understand the meanings of the concepts and infer logical relations, but they do not have the capacity to understand how concepts were related and used to solve complex medical problems for abstraction. The experiment results illustrate there remains considerable room for improvement, especially in utilizing encoded knowledge to perform effective information integration and reasoning (e.g. multi-hop reasoning), pointing toward the avenue of future research in developing NLP models for diagnostic reasoning. 

We showed results of three different T5 in-domain variants training on two different knowledge sources, PubMed (\textsc{SciFive}) and UMLS(T5-\textsc{Defs}, T5-\textsc{RelPaths}). 
\textsc{SciFive} was trained on 32M PubMed abstracts and full-text articles whereas T5-\textsc{Defs} was trained on 515k sentences and paragraphs of definitional sentences, and T5-\textsc{RelPaths} was trained on 582k lines of paths from ``Disease and Symptoms" concept relational graph. T5 continually trained on UMLS data achieved competing performance to \textsc{SciFive} with a corpus that was 3\% the size of \textsc{SciFive}. The smaller corpus of data from UMLS with similar performance to the larger \textsc{SciFive} model demonstrates the importance of a high-quality text corpus. The UMLS model contained both definitional sentences from UMLS that were contextualized semantic representations and concept relation paths that were the knowledge paths for medical concept relations. UMLS is a valuable and free source containing over 127 semantic types and 9 million concept relations. Further work is needed to investigate how to integrate the UMLS knowledge sources for medical knowledge representation.

The gap between the T5-\textsc{Vanilla} and T5 DAPT models decreased as the scale of model parameters increased. Compared to all DAPT models, T5-Large-\textsc{Vanilla} had slightly better performance, indicating some benefit by increasing the network capacity from 220M parameters to 770M parameters. However, the cost and memory requirements for running T5-Large were expensive for minimal gains. While a potential solution to avoid memory restriction is gradient accumulation, the gradient accumulative update took longer and, ultimately, cost more GPU working hours with a larger carbon footprint. One valuable future direction is to distill smaller and more efficient models from large models. In many instances, T5-L only provided marginal gains over T5-B and a more parsimonious model may be the more pragmatic approach for bedside application. Leveraging knowledge sources like UMLS to provide dense, high-quality data, pruning and distillation methods \cite{sanh2019}, as well as other approaches to reduce computing resources are important considerations prior to deploying models at the bedside for clinical use. 



\section{Limitations and Future Directions}\label{limitation} 

Some tasks in DR. BENCH were not initially designed for language generation and may also be solved as a classification task. Half of the tasks were proposed as classification tasks in their original publications (MedNLI, MedQA, AP). The motivation to formulate the tasks as generative was to provide a single configuration and reduce barriers to access and dissemination across multiple datasets so researchers may focus their efforts on the science of diagnostic reasoning. A benchmark for generative models will also leverage the recent advances in large language models (LLM), and allow for the inclusion of future sequence generation tasks as the field continues to grow and evolve.  

The low performance on EmrQA, MedQA and Summ suggests the T5 models have limited capacity in understanding clinical text and inferring relations between concepts, despite best efforts with domain adaptive training. Other approaches like chain-of-thought prompting were recently proposed as a new paradigm of zero-shot learning to invoke LLM's reasoning ability~\cite{wei2022, chowdhery2022}. Using a natural language prompt to illustrate the intermediate steps of complex reasoning has been shown to provide performance gains on arithmetic, commonsense, and symbolic reasoning tasks. 
A separate effort is needed with medical experts to obtain effective chain-of-thought prompts for clinical diagnostic reasoning and this was outside the scope of our work. Knowledge graph is another field that has shown promise~\cite{yasunaga2021, yasunaga2022}. The effort in building DR. BENCH was to organize the benchmarks for useful generative tasks using a single framework to support researchers working in the field of computerized diagnostic decision support augmented by NLP. We leave the establishment of state-of-the-art models with alternative approaches as future directions of DR. BENCH and provide a set of baseline models for comparison.

Another limitation in DR. BENCH is the evaluation metrics and their limitations in capturing the semantics of medicine. The measures of accuracy, macro F1 score, and ROUGE-L employed in our benchmark may be limited in accurately reflecting the many ways the same diagnosis can be generated differently, especially with acronyms and abbreviations that are unique to medicine~\cite{Mrabet2020}. Metrics of redundancy, knowledge representation, and medical context surface the need for human evaluation to overcome the limitation of statistical measures that may not be adequate to assess the reliability of a system prior to real-world testing.


\section{Conclusion}\label{sec13}

Building cNLP systems to perform clinical diagnostic reasoning is a key building block for developing the next generation of NLP-based clinical decision support tools. DR.BENCH has a fine-tuning and multi-tasking setup similar to the GLUE and SQuAD benchmarks in general domain NLP~\cite{wang2019,Rajpurkar2016}, and BLURB in biomedical NLP~\cite{gu2021}, but stands out with a distinct focus and the first cNLP benchmark on promoting clinical diagnostic reasoning. We framed all the tasks as sequence generation tasks, as they served the purpose of evaluating the cNLP models that are designed for computerized diagnostic decision support systems. Researchers could evaluate their generative systems pre-trained on different knowledge sources to examine the progress of clinical diagnostic reasoning, as well as conduct multi-task training using DR.BENCH. We encourage future research to utilize DR.BENCH and shift the focus of cNLP models development from information extraction to complex clinical reasoning. This way, the gap between cNLP models and bedside clinical applications could be filled. As part of the contribution, DR.BENCH pipeline is open-source and released in a GitLab repository, and will be further developed as a leaderboard for the clinical NLP community.

\appendix
\section*{Data Availability}
 ``MIMIC-III' is available at PhysioNet (\url{https://physionet.org/content/mimiciii-demo/1.4/}).
 ``MedNLI'' is also hosted by PhysioNet (\url{https://physionet.org/content/mednli/1.0.0/}). ``emrQA'' and ``AP'' are available at N2C2 (\url{https://n2c2.dbmi.hms.harvard.edu/}). ``MedQA'' could be downloaded from the github repository (\url{https://github.com/jind11/MedQA}). Data for ``SOAP Labeling'' and ``Summ'' are available at PhysioNet (\url{https://www.physionet.org/content/task-1-3-soap-note-tag/1.0.0/}).  

\section*{Code Availability} 
The codes of running all experiments mentioned in this work are available at \url{https://git.doit.wisc.edu/smph-public/dom/uw-icu-data-science-lab-public/drbench}.

\section*{Author Contributions}
Y.G, D.D, T.M, M.C, and M.A conceived and planned the experiments. Y.G, J.C, B.S, and M.A prepared codes and carried out the experiments. Y.G, D.D, T.M, M.C, and M.A contributed to the interpretation of the results. Y.G and M.A took the lead in writing the manuscript. All authors provided critical feedback and helped shape the research, analysis and manuscript. 

\section*{Competing Interest}
No competing interest is declared. 

\section*{Declarations}
The research data used in this work is only available through PhysioNet and the original publication. Data Use Agreement (DUA) is required for MIMIC-III based dataset. We do not claim authorship over the dataset.  

\section*{Funding}
The work was supported by NIH/NIDA grant number R01DA051464 (to MA), NIH/NIGM grant number R01HL157262 (to MMC), NIH/NLM grant numbers R01LM012793 (to TIM), NIH/NLM grant number R01LM010090 (to DD). The content is solely the responsibility of the authors and does not necessarily represent the official views of the National Library Of Medicine or the National Institutes of Health.


\end{document}